# Differences between Industrial Models of Autonomy and Systemic Models of Autonomy

*Aleksander Lodwich*
*aleksander[at]lodwich.net*

**Abstract** – This paper discusses the idea of levels of autonomy of systems – be this technical or organic – and compares the insights with models employed by industries used to describe maturity and capability of their products.

## 1  Introduction

The quantification of system autonomy is deemed desirable as measure of performance of products. The term *autonomous* proposes to customers the idea that they will spend less time and effort on monitoring a system or fixing problems observed during its operation. Additional benefits of autonomy are considered the ability to keep a system operational even if communication channels between operator and device fail.

However, I observe a new trend to redeclare projects formerly titled as autonomous as automatic again. It is also observed by others that efforts made towards definition of "levels of autonomy" do not contribute effectively to solutions and are hence discouraged for practical engineering parties. For example, the U. S. Department of Defense warns to invest effort into autonomy models because they do not contribute to improved armed forces performance [1]. Instead, the DoD concludes that creating optimized systems for selected scenarios has proven cheaper and more satisfactory and that this more traditional approach does not preclude creation of systems with resilient features or some temporary degree of self-governance.

This paper will delve into the question whether calling systems "autonomous" is useful anymore. Furthermore I explore the possibility that industry has tried to define the levels along the wrong lines of thought.

Down this paper I will propose a different model of systems autonomy that is suited to explain the drift away from attributing technical products as autonomous and to call them automatic again. I will also discuss the concept that industry will never be motivated to produce autonomous systems but only highly automatized products.

## 2  Concept of Autonomy

Most of scientific literature attempts to define autonomy in context of particular applications. For example, Beer [2] made a survey of autonomy definitions for the area of robot human interactions. Other domains, such as mathematics, ethics, education or economics define their own concepts of autonomy. Because the definitions are expressed in domain-specific terms, those definitions are not to be considered systemic, i.e. they cannot be applied to arbitrary systems, contexts or domains.

Beer's collection of definitions is most relevant for engineers as it deals with technical products. According to his findings most authors cited by Beers describe autonomy as the capability to

- not require permanent control
- refine goals
- deal with changing environments
- perform cognitive tasks
- act rationally
  (regarding defined goals)

Such functions are often organized in "levels of autonomy". Beers relates to the ALFUS framework [3] as an example for this. Today, defining "general levels" of autonomy is perceived as inadequate as this would imply some general level of autonomy across the many functional aspects of technical systems. However, current (robotic) appliances often demonstrate autonomy in particular areas only, as for example an ability to fix a physical impasse, correct sensor noise, update route or to simplify human surveillance and monitoring activities by selective data presentation. Such systems can prove extremely flexible at higher levels of abstraction of activities but could prove quite helpless at dealing with basic behavioral challenges and vice versa.

Beer's competing model organized around the sence-plan-act cycle tries to explain various degrees (in-)abilities of robotic appliances and his work is just an example of a series of attempts to abandon simplistic "levels of autonomy" models for the robotic domain and other technical areas, such as pure software agents.

Other researchers do not try to explain autonomy in terms of appliance's capabilities. Ziemke [4] surveyed and discussed autonomy concepts from biology, among them concepts of autopoesis (Varela, Maturana), recursive self-maintenance (Bickhard), relatively recent work of Christensen and Hooker or older work of Uexküll.

I brace out deeper treatment of those ideas here but it is important to highlight the conclusions drawn by Ziemke in that autonomy, as is observed in organic systems, is different from "artificial autonomy" that engineers strive for. One could differentiate between "natural autonomy" and "artificial autonomy". Figure 1 shows a basic concept of this idea.

Despite that authors researching properties of "natural autonomy" occasionally mention or imply "higher levels" of autonomy - among them one could be responsible for creation of conscious experience - I am not aware of any "standard" model to characterize the amount of autonomy of a system from this line of research and will hence give a try in this paper.

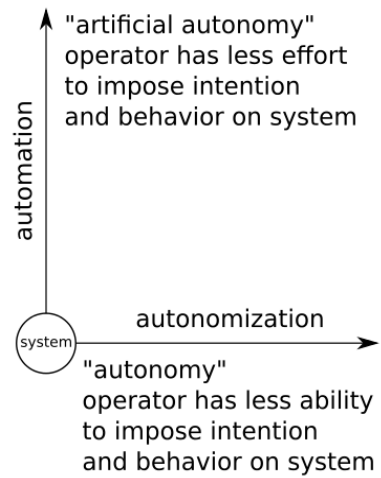

*Figure 1: natural and artificial autonomy strive for different goals*

In general, I will follow the systemic autonomy concepts of Bickhard, Christensen and Hooker who explain autonomy of a system as the ability to maintain its structure in a dynamically changing environment. This structure needs not be necessarily closed – contrary to autopoietic ideas presented by Varela and Maturana earlier on but it should be intuitively clear that enforcement of structure is technically easier in setups with guarded boundaries. Bourgine and Stewart treated the question of self-maintenance and autopoiesis in closed and open systems in more detail [5]. Their extended definition of autopoietic properties relies on the idea of a process reprint over time. Ideally, such autonomous sets of processes show immune to attempts to control their function via external variables. If it was possible to define different degrees of influence on such autonomous sets then it could be possible to define a relatively robust systemic model of levels of (natural) autonomy in terms of how well the system can immunize itself against a certain type of attempt to alter its internal function.

It is very interesting to apply this concept of autonomy to dead things such as a rock. Indeed, a rock will preserve its structure quite a bit of time and will withstand quite a bit of external influence. At first it appears alienating to call a piece of rock "autonomous" but at a second glance it could be quite convincing: The rock consists of billions of atoms communicating with each other by electro-mechanical forces maintaining and correcting the structure of the rock in all conditions. It is the characteristic of a systemic theory of autonomy that it is applicable to all systems but it also means that not all autonomous systems are necessarily all that exciting to analyze or that they are equipped with sophisticated cognitive capabilities.

However, question of achieving autonomy is posed mostly in context of life-like things. Barandiaran and Moreno clustered various types of processes in life-like systems in a (mostly) nested manner in order to explain self-maintenance in changing environments [6]. This would be the closest to a "levels of autonomy model" that I am aware of but since it emphasizes the idea of "nested control" and because it covers internal and external aspects of maintenance, it is difficult to show how a life-like system could not have any of the process types. In contrast, the idea of levels is that elevated levels are optional.

None of the authors seems to treat autonomy from a reproductive point of view. Terminology like *self-maintenance* and *autonomy* suggest properties of instantiated singular objects but the idea can be extended to classes of objects as well, opening the opportunity to think about (self-)maintenance of systems and structure not only over time but also over space. In that extended view a system (class) can gain autonomy by reprinting unaffected instances if it cannot self-maintain anymore. One could call this as some kind of "teleporting" self-maintenance but reproduction is the more common term. Example of this type of autonomous entity could be the computer virus spawning multiple instances of itself. It would prove very hard to kill. Even if individual instances were eradicated, it could still manage to follow a specific agenda (such as encrypting disk files). In biology the analogy would be a species of organisms which would become autonomous by implementing reproductive features.

An extreme view of autonomous systems would fully concentrate on reproduction and would take a space-time approach. It would simplify to explain how systems can elastically and dynamically adapt their forth-existence strategy between reprinting in the same place (reproduction over time) or reprinting in a different place (reproduction over space). A hypothetical autonomous system (e.g. some kind of organism), can follow a strategy to evade influence (control amount of inference from external variables), to harden its protections (lower sensitivity to external variables), increase its healing capabilities (faster replacement of compromised blocks which were impacted by external variables using newly replicated blocks) or replicate altogether its state into a different place where certain variables (e.g. predator's states) do not affect it.

Moreover, autonomy can be temporary and contextual. I call this *relativistic autonomy*. For example, a system could have an unlocking mechanism that would allow insertion of dependencies to "external" variables. After insertion, such variable or system of variables could be considered as "part of" the target autonomous system. Autonomous systems with self-discovery capabilities will indeed detect the insertion and reorganize its boundaries (if they exist) in order to protect the attached variable or system of variables.

Such mechanisms require a long access key (in terms of bit of information or units of energy) by the environment in order not to inter-fere with regular operation of an autonomous system. This is expressed as a resistor in figure 2. If defended off systems can access variables or systems of variables against which the autonomous system shows little defense then the system can either fall under control of a foreign system or, if it is learning by experience, has to increase autonomy against such channels. Sophisticated systems will therefore manage their "trust" and "security" actively, in order to retain autonomy in other areas.

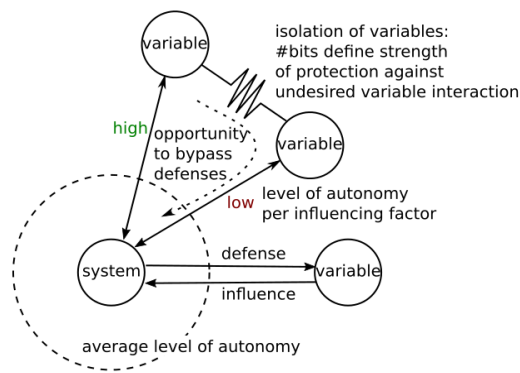

*Figure 2: Relativistic autonomy is a concept similar to saying "is independent of" other influencing systems and their variables but it is not equivocal with it.*

Relativistic autonomy introduces the concept that property of autonomy can be the result of an angle at which the environment is interacting with a system. This again brings up security topics when designing products.

## 3   Levels of System Autonomy

It is quite natural to assume that attaining higher levels of autonomy for a system (set of autonomous processes) would require improved capabilities of behavior control (cognitive capabilities). However, this could be a falsehood. Let us imagine a scenario in which a huge armored animal with very basic cognitive capabilities is pursuing its business when becoming target of a group of very intelligent, smaller and weaker creatures. It is not difficult to imagine that the heavily armored animal can retaliate, repel an intelligently orchestrated attack and defeat the attackers. In that case it would prove unaffected by systems of higher mental capabilities. However, it could become victim of a cognitively less sophisticated predator with bigger mass and force. In that case the autonomous system would fall under external control (and the animal would be eaten).

Defining critical points on which a system would fall short of autonomy or lose it, is interesting for defining levels of (systemic) autonomy. Because cognitive capabilities clearly compensate for strength, speed, energy and vice versa, it is not possible to define levels of systems autonomy in pure cognitive terms such as "can reason ahead of time" or "can detect invariants in sensory stimuli". Nevertheless, as we will see, coping with very aggressive scenarios requires types of internal system capabilities that are familiar with what we know of cognitive agents. Moreover, the next proposed model does not suggest that systems in higher levels of autonomy can automatically take control of systems with lower levels of autonomy. For example, a human could move a rock but not crush it. In this sense a higher level autonomy system is not necessarily capable to compromise a lower autonomy level system. However, a crushed rock will never return to its original structure while a broken leg will heal. Levels of autonomy are useful to give a prognosis how a system will behave after being structurally violated.

In order to attain a systemic levels description I will try to define them in terms of systems and how their reproduction policies have to improve. I will introduce a new level to the model each time whenever environmental aggressiveness raises to a point where qualitative enhancements of the system policy are mandatory. Since autonomy is about structure, violation of this structure and later restoration of it, a key concept to understanding the various levels of autonomy is coming through configuration theory. Such configuration theory is organized of plants (resources to be configured), retreat storages (memory of configurations), re-configuration mechanisms and the information transfer speeds between plants, retreat storages and re-configuration facilities. An important concept in configuration theory is the cost of re-configuration that is a key modulator to expression of system's other affordances. As a consequence we will see more and more cognitive features slip into the level descriptions but since the definitions are systemic, the levels do not unavoidably demand the equivalent of a cognitive apparatus in the system. For example, a sponge will react adaptively to external forces and return to its original state without anything comparable to a dedicated nervous system. Despite lacking units of behavior control, a sponge will "remember" its original form and will be able to "restore" it at a finite speed. That's the "systemic" property of the approach. However, if read carefully, the following level definitions should translate relatively easily into known capabilities of central nervous systems or cognitive control units.

The following level descriptions represent each level as is shown in figure 4.

### 3.1 Level 0 – Not autonomous

Level 0: At this level a system is not autonomous. It is characterized by permanent process of structure deterioration. Deterioration is caused by the faintest external influences. A collection of stones forming a circle on a table is an example of a system that is not autonomous. Small vibrations or a push of the wind will destroy the formation. The example system has no policy of maintenance, does not consume energy or information for this purpose and has no means to remember its original configuration. However, the classification as level 0 may depend on the exact system boundaries. For example, a vacuum cleaner may not be autonomous (level 0) but if you draw the system boundary around human users and vacuum cleaners then we detect a symbiotic relationship among those systems that warrants autonomy to its non-autonomous parts (Figure 3). In the suggested scenario vacuum cleaners can defy deterioration and attacks on it by exposing certain type of utility to users and by being reprinted by them. The wide success and presence of vacuum cleaners in every household should clearly explain how the system guarantees its existence by reprinting itself over time and over space. This would elevate the scenario to level 1 autonomy.

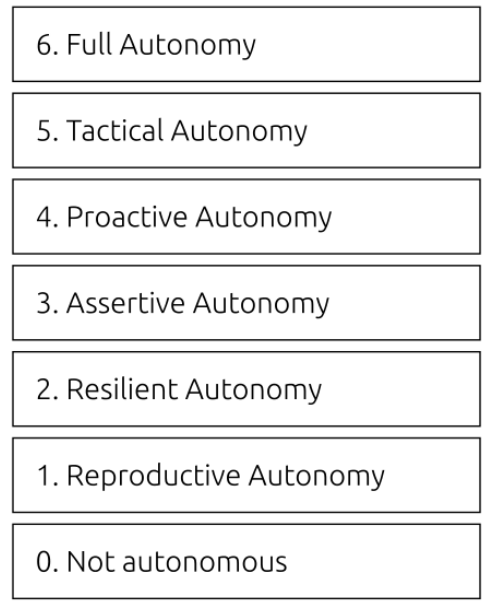

*Figure 4: Proposed levels of systemic autonomy*

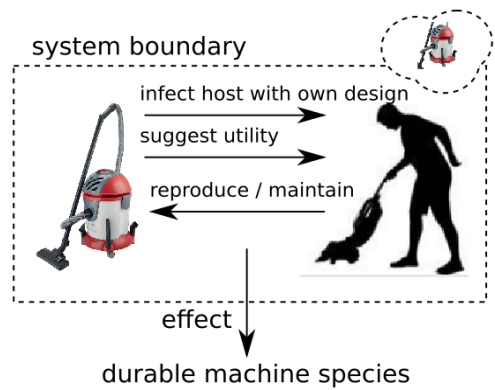

*Figure 3: The vacuum cleaner species appears autonomous*

## 3.2 Level 1 – Reproductive Autonomy

Level 1: At that level the environment is following static rules of behavior which must be immunized against. A static environment is not without motion or flow but the dynamics are characterized by relatively stable higher derivatives close to zero. In such an environment a static policy can be sufficient in order to maintain system's *conditions of existence*. Similarly to the static environment, a static policy is not necessarily without behavior. Generation of a pulse-width signal to represent 75% or power (cf. figure 5) is clearly involving changes of state but is nonetheless a static behavior policy and hence static. Such policies need not necessarily be simple: Finite automatons for controlling system behavior with billions of states and transitions are still static in the here necessary sense.

A more concrete example of a level 1 autonomous system would be a fish that is optimized for a certain water depth and water speeds (cf. figure 6). It is irrelevant that swimming is a complex process: As long as it reproduces in a certain area and implements a certain depth holding mechanisms and maintains a constant swimming speed, it will survive in the stream of water and, in fact, could be quite successful in doing so. Please note that in the case of the imaginary fish the systemic policy consists of moving muscles with a certain force and speed. The success of this policy lies in the equilibrium of the forces of water and forces of propulsion and a minimum satisfaction of energy supply from the food. This success does not mainly rely on sophisticated behavioral features but on energetic and mechanical properties of the system. If the system's policy told to develop weaker muscles then the fish could not survive in the optimal water depth.

However, if the flow of water changes – for example by changing the layering of flows – then this fish species is highly endangered because the ability to adapt to such changes is a feature of level 2 autonomy.

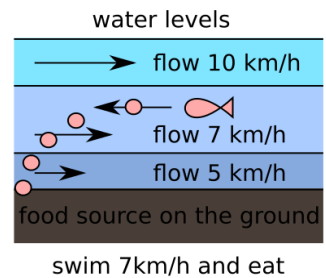

*Figure 6: An imaginary fish swims at a certain speed and depth, eats and breeds there. In a static environment this is good enough.*

I would like to side-remark that creation of life (autonomous compounds of matter) will probably necessarily go through this level of autonomy first. One can shorten it down to "it exists because it works" and requires generous environments – some kind of primordial soup of benign conditions. As a consequence, autopoietic systems developing further into less benign environments will concentrate on protecting its core conditions by creating adapter mantles and protective boundaries (cf. figure 7).

Unless a system is some kind of rock, it will suffer from deterioration and will require an active part on its own in order to restore its structure. This will translate a basic "it exists because it works" into numerous drivers necessary to control the more complex protective adapter mantles. Ziemke relates to work of Damasio in [4] and draws a picture in order to explain this relation as is shown in figure 8. However, he leaves it open what the layers mean and where they could come from. I pro-pose to align them with autonomy level adapter mantles.

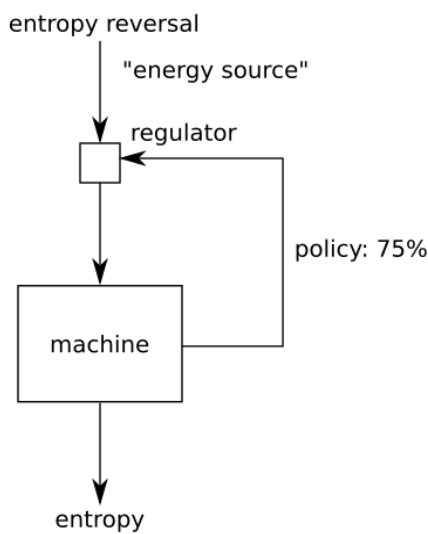

*Figure 5: A static strategy to reprint the system over time despite machine deterioration. This is an example for autonomy of level 1.*

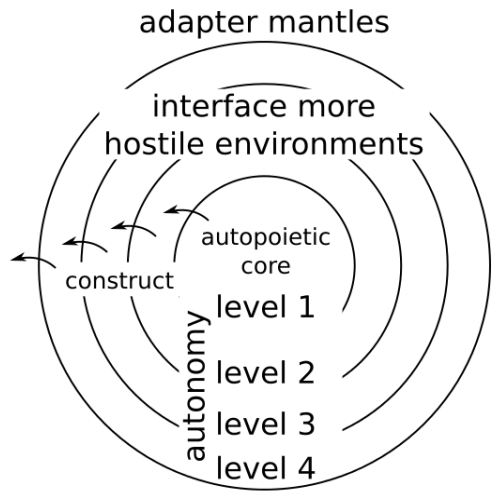

Figure 7: Radical approaches to protection of core autonomy result in creation of sophisticated "environment adapters".

Since such adapters can be functionally layered in their own right, the amount of possible drives to control all the functions is indeed potentially infinite. Functional layers are very often used as a model of hierarchical control in the environment. Especially systems with learning capabilities will generously expand such functional layer architectures. Therefore, it is very difficult to predict the exact expression of drivers in systems which are allowed to grow into a certain niche. This is also an everyday experience that people living in different sub-societies also feel different about great many things and constantly look for new concepts to identify and grasp the differences.

### 3.3  Level 2 – Resilient Autonomy

Level 2: In level two autonomy is protected by allowing the system to adapt to occasional changes in the environment by overwriting the static policy with a new equivalent static policy. For example, if the energy source weakens in figure 5 then system can crank up the power rate of regulator. In contrast to level 1 policies, level 2 are unbounded. If we bounded them then we could convert them into level 1 policies for practical reasons but it is important to understand the loss of flexibility. The most generic and unbounded means to alter policies is to combine random

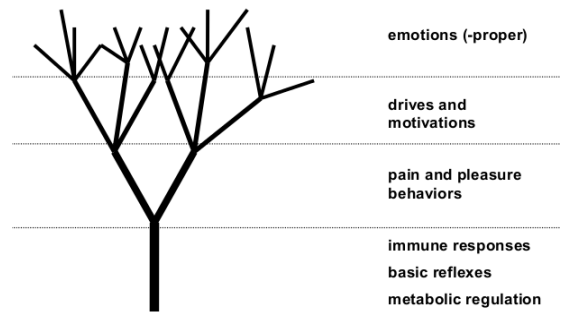

Figure 8: Ziemke draws a causation tree of emotions down to a common root originating in basic organism functions such as metabolic regulation.

modifications of the policy with adaptive resonance.

Level 2 adaptation features are lossy: Once a system adapted to a new situation it has lost the old ability. Think of a closed-loop controller in cybernetics. It can be designed with an outside adapter to compensate for slow wear out of system. However, if system is maintained and wear is removed then it has to slowly adapt back into its original parameters. In real life the controller would be simply reset to original values but in autonomous systems such resets are limited from the outside. A system would have to learn to reset itself in order to prevent wasteful adaptations.

Environments, in which processes can take place faster than the system can adapt to, can exploit this fact in order to gain control over the autonomous system. Therefore, autonomous systems in level 2 would be best advised to become constantly faster. However, that is a feature of level 3.

In figure 8 the bottom layer of drivers is corresponding to level 2 autonomy in my model.

### 3.4  Level 3 – Assertive Autonomy

Level 3: In that level of autonomy a system can protect itself against environments growing faster in speed.

Faster, in terms of configuration theory, will require a minimized role of adaptation and increased use of configuration templates to the task. This will indeed require the retreat storage (memory) for the first time and will de-mand triggers to start a re-configuration process. Ideally, the system can organize stored configurations and implementation facilities that it develops by the amount of shared information in order to minimize internal bandwidth requirements which are limiting the maximum re-configuration speed.

If system is experiencing new states it can try to remember them either as discrete chunks or probability distributions. Triggers in level 3 mantle adapters must somehow translate into violations of more fundamental conditions. This gives raise to regulatory signals which trigger a sudden reconfiguration of policy. For example, people suddenly stop working and go to kitchen to fetch some food. This sudden changes of operation are key improvements to overall efficiency of the system. The system can execute highly efficient programs sequentially and minimize effects of the no-free-lunch theorem stating that no (static) policy will ever be good in better than a fraction of possible cases.

Important concept of level 3 autonomy is that of growing complexity. Because of the technical demand to switch faster there is a constant growth of internal dimensionality of the system ("grows in experience"). Over time, it adds policy implementations and triggers to switch them.

In figure 8 the layer of pains and pleasures is corresponding to level 3 autonomy in my model. Various kinds of pain and pleasures are important triggers to switch major groups of policies.

### 3.5 Level 4 – Proactive Autonomy

In level 4 the autonomous system will try to synchronize with future events for which there is no precursor or which suffer from *computational irreducibility* [7]. Environments which are free to act unpredictably (e.g. environment is tactically constructive) can take advantage of this in order to gain control over the autonomous system. The system must develop policies and triggers for them before they actually occur. This will require several basic things: The system must not only implement policies to interact with the environment but also must model the environment itself one way or the other. It can exploit these models in order to train a new policy without the associated risks. For example, if a person sees a door with a knob on the inside then it can quickly imagine how it would enter the room, get locked and not be able to get out again. This would alter its strategy to use the room and could be interpreted as logical reasoning. Readers interested in ideas of system synchronization and reasoning through simulation can refer to survey-type articles made by Pezzulo [8] and Grush [9].

In order to perform such *synchronization with the future* – as Pezzulo puts it – systems will have to reserve and dedicate parts of its own plant to simulating the virtual twin of the environment. In fact the simulations must include some kind of models related to the system in order to be useful. Creation of a self-model appears necessary to do this successfully or the simulation would directly interact with the real plant in a disturbing way.

Since simulating all aspects of the environment appears technically infeasible even in theoretical terms, the autonomous system must apply trigger/selection and re-configuration capabilities obtained in previous development stage in order to set up a proper simulation environment.

One of the most important problems for systems is to compensate for reaction time (dead time) in the environment. Many activities must start long time before any useful feedback or trigger can fire. In order to achieve this, modeling of the environment using clocks is critical. A typical example of synchronization is the use of calendars in order to optimize farming activities. Farmers have to work for months before seeing any crop to harvest.

Since simulations can follow various lines of development, a system must be able to select one of the choices and schedule it for execution. Readers interested in how systems could model and exploit such models for decision making are referred to work of Paul Werbos [10][11].

In figure 8 the layer of motives is corresponding to level 4 autonomy in my model.

Motives get created & disappear along the course of temporal planning and are strongly associated with mental simulations which they should guide.

### 3.6  Level 5 – Tactical Autonomy

Life is full of choices. What is to be said by this is that an autonomous system will face an environment where it cannot prevent losses and hence must decide between critical and expendable resources. This is even more pressing as the environment is full of equivalent objects but the allocation of losses among them is not irrelevant for the system making the decision.

If a system cannot detect and properly handle meta-patterns of its models then it can be overwhelmed by an environment creating functionally equivalent permutations. This sounds very much like autism. Possibly, autism is a consequence of defects when children develop features of level 5 autonomy protecting adapter mantles. It is speculated that vaccination with mercury and aluminum, which are very toxic to nerves, is causing this modern day mass phenomenon. In any case, if a system development failure leads to the inability to detect invariants, intentional creation of permutations would disturb all triggers, models and simulations of the level 4 agent and hence would allow external systems to fool it under own control.

Since the environment is full of repeating patterns and variations of it, it is critical for a system to expand models to manage them in a taxonomic way. This will also allow to reuse functionality across various new situations. Once this is done, same concepts will pave any modeled scenario and the agent must have means to specialize (de-generalize) his scenarios in a way that would satisfy lower-level drivers: For example a fighter and his adversary are both humans but in a fight situation a fighter cannot optimize the situation for both. Only one instance is eligible for motive satisfaction. However, through the generalization of concepts the "fighting function" has become independent of many properties of the adversary.

It is clear that gaining on capabilities in level 5 a system can not only avoid traps or synchronize with future events but also avoid simulations by accessing equivalent "wrapped up simulations" in form of conceptualized short stories which can be reapplied to each object individually and arrive at stable conclusions irrespectively of many seemingly never repeating situations.

So, the main new technical feature for level 5 agents are to reuse chunked model components multiple times, to organize them hierarchically and to skew treatment of entities of same class depending on the side of boundaries they are on (inside/outside). These are indeed fundamental features for making game theoretic considerations. That is why I chose to call this level of autonomy *tactical*. Again, even if this begs for some cognitive nervous system feature, this is not necessarily so. A group of military generals can use a pool of toys moved on a table in order to simulate and verify certain strategies, some of them at various levels of abstractions. If they are successful in this endeavor then they can contribute to the autonomy of the protected state. In fact, historically, game theory is output of military analysis activities.

Admittedly, the main obstacle to achieving this kind of generic knowledge base lies in making all models interoperable and to my knowledge it is not fully resolved. It is out of scope of this paper to treat integratability, composability or universal knowledge representation approaches but mastering them is key challenge to fully harness level 5 autonomy.

In figure 8 the layer of emotions is corresponding to level 5 autonomy in my model. Emotions towards concrete entities help to amplify non-universal, self-centered policies which are necessary in order to stabilize own structure in face of many similar, ephemerally changing structures surrounding the autonomous system.

### 3.7  Level 6 – Full Autonomy

How can the level of difficulty can be raised even higher? Well, even if systems can exploit their resources better by improving their internal organization, they still can suffer from significant dependencies that can only be alleviated with greater size of the autonomous system. Level 6 will require effective communication with external systems in order to distribute risks and effects of major variables. A species capable of level 6 autonomy is equipped with features simplifying creation of larger level 1 entities of which they are part and which participate from the strength of the clusters. This extends the effective control over influencing factors beyond immediate reach. This is a democratic view. Striving for full autonomy is often much less egalitarian. In monarchic structures only a single unit of a group could claim full autonomy at the cost of reducing autonomy of all other units. Such reductions are only acceptable by the units if their cost-benefit calculus is on the positive; meaning that gaining higher level of autonomy in that context would (re-)introduce hard to control risks.

The process of organizing groups of level 6 units is recursive. New groups become created spontaneously on the "it exists because it works" basis and through a series of more or less painful events the unit is upgraded and mantled according to the schema in figure 7. Figure 9 shows this recursive process.

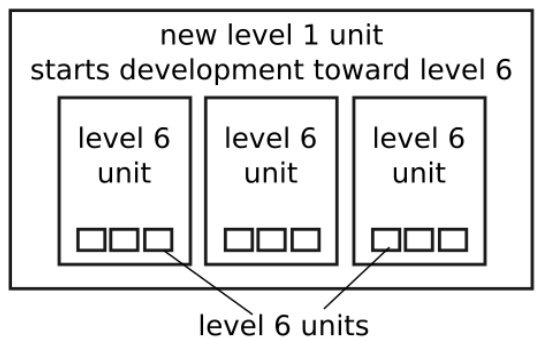

*Figure 9: Autonomous systems form larger autonomous systems which will mature until all resources are exhausted.*

# 4 Conclusions for System Designers

The industry's use of the term "autonomy" is slowly replaced by the term "automatic". For example, the SAE J3016 standard is defining six automation levels for what we call "autonomous vehicles" today. Automation is correctly replacing the term "autonomy" step by step in the industry because technical products shall not be autonomous. Industries are not interested in creation of products which can escape control of human operators.

However, perception of autonomy can depend on perspective: A highly automated weapons system is perceived as "autonomous" by the opponent (who cannot impose control over the system) but is perceived as "automatic" by the weapons operator. Such perspectives are the result of relativistic autonomy which I have explained.

For practical purposes, the question of a/a-perspective is better treated under the question of secure control and the questions of potential loss of control from poorly implemented security mechanisms. A highly automated system could become autonomous in situations where authorized control facilities stop working (operator gets "locked out") or it could implicate such state when indeed the system has been brought under foreign control. In all cases this would be considered undesirable.

What is undesired for products is desired in self-maintaining systems. In this paper I have proposed a six level model of "natural" autonomy that is organized by the amount of aggressiveness of a system's environment. The riskier, the more manipulative and deceptive the environment is, the higher are the requirements toward the autonomous system in order not to fall under foreign control (and risk disintegration). This is at odds with industrial autonomy models where the relationship between operator and product is defined and not challenged. Operator should not lose control of the system but have less effort to impose his will on it – so to say. Some researchers call this "artificial autonomy". Even the most automatic products on the market like the Google car are not reaching levels of autonomy beyond level 2. There seems to exist no good reason why Google car should ever be equipped with mechanisms for increasing its internal complexity in order to make it level 3 unit.

However, systems with low or absent autonomy can participate in higher level autonomy systems by being embedded in a new systemic context. This will give external observers the impression that these systems attained higher level autonomy. This can be an important instrument to create pseudo-autonomous products if true autonomy shall be avoided. Pseudo-autonomy is best explained with externalized reproduction mechanisms. I have briefly discussed this matter on the vacuum cleaner example.